\documentclass[conference]{IEEEtran}
\IEEEoverridecommandlockouts
\usepackage{cite}
\usepackage{amsmath,amssymb,amsfonts}
\usepackage{algorithmic}
\usepackage{multirow}
\usepackage[table,xcdraw]{xcolor}
\usepackage{bbding}
\usepackage{hyperref}
\usepackage{float}
\usepackage{graphics}
\usepackage{graphicx}
\usepackage{textcomp}
\usepackage{xcolor}
\hypersetup{colorlinks=True,
            citecolor=black,
            urlcolor=blue,
            linkcolor=red,
            }
\def\BibTeX{{\rm B\kern-.05em{\sc i\kern-.025em b}\kern-.08em
    T\kern-.1667em\lower.7ex\hbox{E}\kern-.125emX}}
\newcommand{\linebreakand}{%
\end{@IEEEauthorhalign}
\hfill\mbox{}\par
\mbox{}\hfill\begin{@IEEEauthorhalign}
}
\makeatother
\begin{document}

\title{Image-text Retrieval via Preserving Main Semantics of Vision
\thanks{*Corresponding author. This work is supported by National Natural Science Foundation of China under grants (No.62272087); Science and Technology Planning Project of Sichuan Province under Grant (No.2023YFG0161).}
}

\author{
\IEEEauthorblockN{1\textsuperscript{st} Xu Zhang}
\IEEEauthorblockA{\textit{School of Computer Science and Engineering} \\
\textit{University of Electronic Science and Technology of China}\\
Chengdu, China\\
zhangxu0963@163.com}
\and
\IEEEauthorblockN{2\textsuperscript{nd} Xinzheng Niu$^*$}
\IEEEauthorblockA{\textit{School of Computer Science and Engineering} \\
\textit{University of Electronic Science and Technology of China}\\
Chengdu, China \\
xinzhengniu@uestc.edu.cn}
\and
\IEEEauthorblockN{3\textsuperscript{rd} Philippe Fournier-Viger}
\IEEEauthorblockA{\textit{College of Computer Science \& Software Engineering} \\
\textit{Shenzhen University}\\
Shenzhen, China \\
philfv@qq.com}
\and
\IEEEauthorblockN{4\textsuperscript{th} Xudong Dai}
\IEEEauthorblockA{\textit{School of Computer Science and Engineering} \\
\textit{University of Electronic Science and Technology of China}\\
Chengdu, China \\
daixudong\_s@163.com}
}

\maketitle
\begin{abstract}
Image-text retrieval is one of the major tasks of cross-modal retrieval. Several approaches for this task map images and texts into a common space to create correspondences between the two modalities.
However, due to the content (semantics) richness of an image, redundant secondary information in an image may cause false matches. To address this issue, this paper presents a semantic optimization approach, implemented as a Visual Semantic Loss (VSL), to assist the model in focusing on an image's main content. This approach is inspired by how people typically annotate the content of an image  by describing its main content.
Thus, we leverage the annotated texts corresponding to an image to assist the model in capturing the main content of the image, reducing the negative impact of secondary content. Extensive experiments on two benchmark datasets (MSCOCO and Flickr30K) demonstrate the superior performance of our method. The code is available at: \href{https://github.com/ZhangXu0963/VSL}{https://github.com/ZhangXu0963/VSL}.
\end{abstract}


\section{Introduction}
Cross-modal retrieval is an emerging task that matches instance from one modality with instance from another to bridge the gap between different modalities. One of the most important cross-model retrieval tasks is image-text retrieval. It requires building a common representation space for images and texts. The key challenge lies in learning the alignment of image and text to accurately measure similarity of image-text pairs. 

In recent years, with advances in image feature extraction and natural language processing, image-text retrieval methods based on inter-modal correspondence have become prominent. Some prior methods ~\cite{SGRAF, IMRAM} learnt image and text in a common embedding space and directly measure the similarity through the cosine similarity metric. Despite  that this can  yield excellent results,  correspondence between contents of the same modality is ignored. To address this issue,  approaches were designed~\cite{MMCA,ILSA} that mine valuable intra-modal information to assist fine-grained feature alignment. Wei $et~al.$~\cite{MMCA} jointly modeled the inter- and intra-modal relationships of regions and words with attention modules, which resulted in a significant performance improvement. Wang $et~al.$~\cite{ILSA} made full use of instance-level intra-modal information to generate more appropriate distributions.
\begin{figure}[t]
    \centering
    \includegraphics[width=8cm, trim=50 160 290 100]{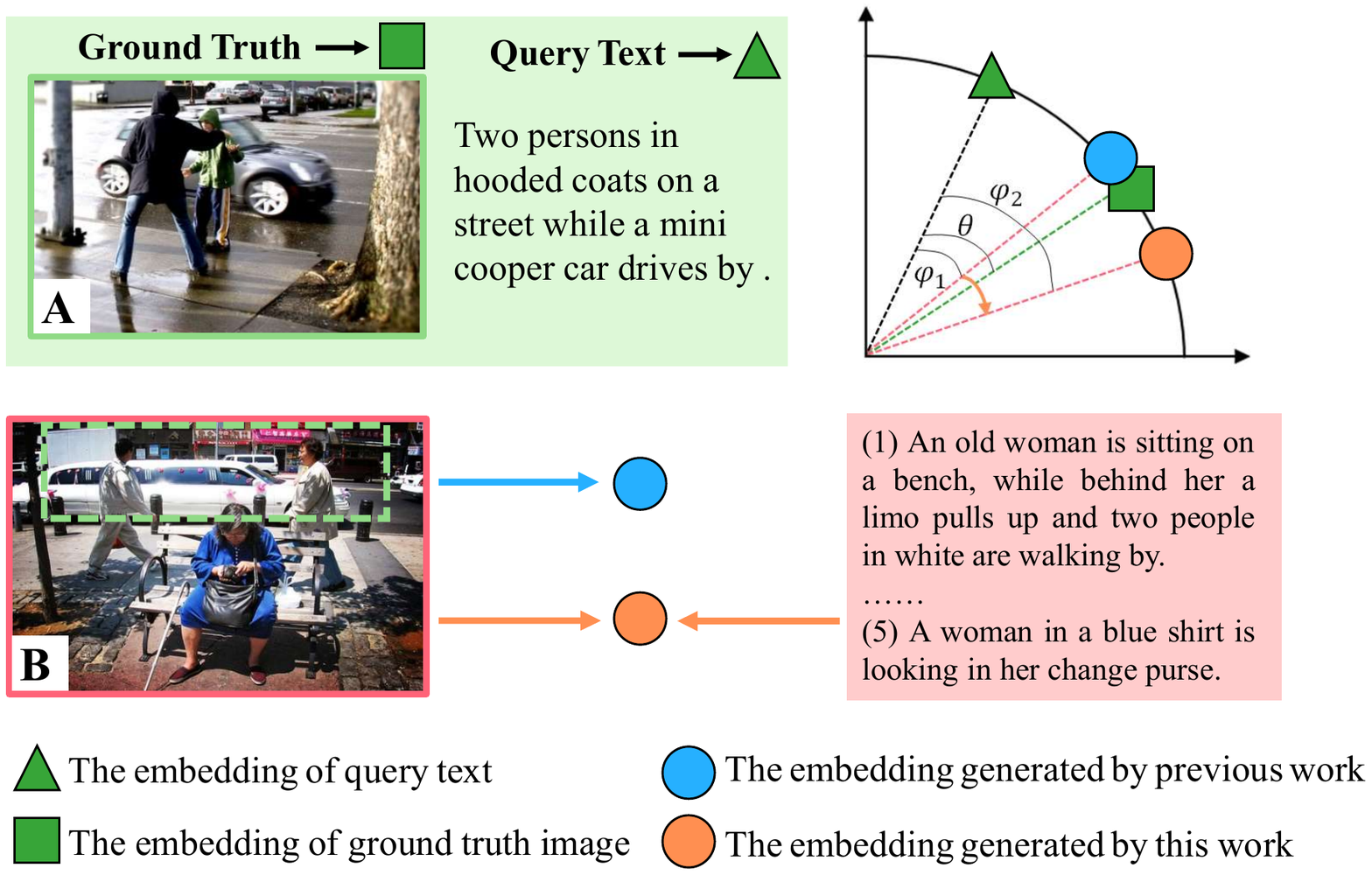}
    \caption{A comparison of previous work with the proposed method. In conventional retrieval, image B (non-ground-truth) is closer to the query than image A (ground-truth), shown as ${\varphi_1<\theta}$. For the reason that the secondary content in the background of image B, shown in the green dotted box, contains the scene described in the query. Differently, the retrieval based on the proposed \textbf{V}isual \textbf{S}emantic \textbf{L}oss (VSL) ensures that the focus is on the main content and push the negative sample (image B) away from the query, represented as ${\varphi_1<\theta<\varphi_2}$.}
    \label{f1}
\end{figure}

However, the aforementioned studies fail to capture the main semantics of images in the presence of secondary content, which can cause incorrect retrieval. Concretely, an image consists of several different semantic regions, from where model can obtain abundant features to describe different contents. However the main semantics of an image are typically only related to a portion of the contents, and the remainder may cause a mismatch with the text. Previous methods~\cite{SGRAF,MMCA,CMHF} do not distinguish well between main and secondary contents in an image. As a result, they often mistakenly match an image and text because the secondary content of the image is consistent with the text, but they ignore that the main image content is irrelevant to the text. As shown in Fig.~\ref{f1}, given ``Two persons in hooded coats on a street while a mini cooper car drives by.'' as a query with image A as ground truth, the baseline~\cite{SGRAF} wrongly match the image B because the model ignores the main content of image B described by annotated texts but only focuses on the minor contents in the background (green dotted box in image B).

To overcome this issue, this paper presents a semantic metric learning approach, which enhances the relevance between positive pairs. Inspired by people that tend to focus on the main content of an image when making annotations, we leverage the annotated texts of an image to assist the model in capturing the main semantics. Annotated texts (five per image in MSCOCO) are regarded as Positive Texts (PTs), which describe the main semantics of an image. These PTs are used to distinguish between main and secondary contents. Furthermore, a novel loss function is designed named \textbf{V}isual \textbf{S}emantic \textbf{L}oss (VSL) to ensure that the model prioritizes matching texts to images whose main content is consistent with the text description as much as possible. Inspired by~\cite{DAA}, the semantic similarity within visual modality is estimated using CIDEr~\cite{CIDEr} among PTs, since high semantic similarity between PTs of two images implies that the main contents of two images are highly similar. Numerous experimental results show that the proposed VSL considerably improves the baseline model, which prove that the approach enhances the semantic representation of the main content of the images.

The main contributions of this paper are summarized as follows. (1) A critical challenge is identified in cross-modal retrieval that a model incorrectly matches texts with images because the primary content of the image is irrelevant to the text, yet secondary content is similar to the text. (2) To address this challenge, a novel metric learning method is proposed, which specifically acts as a Visual Semantic Loss (VSL). The approach uses the annotated texts of an image to assist the model in capturing the main semantics of the image which enhances the alignment of the text with the main content of the image. (3) Quantitative and Qualitative experiments with~\cite{SGRAF} as baseline are conducted on two commonly used datasets ~\cite{COCO,f30k}. The proposed approach yields superior results compared to the state-of-the-art methods.
\section{related work}

\textbf{Cross-modal Retrieval.} Due to heterogeneity among multiple modalities, the widely used alignment method is to map data from different modalities into a common space and measure their similarity. Prior methods can be grouped into two categories, that is methods using intra-modal information and those that do not. This paper focus on the former one. Kipf $et~al.$~\cite{Semi} introduced Graph Convolutional Network (GCN) to learn the relations between image regions within visual modality to generate a visual representation capturing key concepts of a scene. Then, Li $et~al.$~\cite{VSRN} extended the GCN by proposing a fusion mechanism for local and global intra-modal features. To capture high-level interactions between regions and words, Xu $et~al.$~\cite{CMHF} fused multimodal features with inter- and intra-modal relations. In another study, Wang $et~al.$~\cite{C3CMR} proposed intra-modal contrastive learning to mine neglected relationships, and consequently enhance the diversity of both positive and negative instances. Li $et~al.$~\cite{ALBEF} designed a contrastive loss to align the image and text representation in the large-scale dataset. Moreover, a graph structure was used by~\cite{ABGR} to represent each modality by graph and embed the graph-structured information in a common space. 

\noindent\textbf{Metric Learning.} This kind of method measures the similarity between modalities using distance metrics represented as loss functions. The widely used loss function for cross-modal retrieval task is the triplet loss~\cite{FaceNet}. It was used to measure the similarity among a triplet to guide positive samples closer to each other and farther from negative sample. A series of studies~\cite{N-pair,angle,ANG} developed triplet loss in different aspects. Sohn $et~al.$~\cite{N-pair} proposed N-pair loss which keeps the negatives away in each triplet. Wang $et~al.$~\cite{angle} designed an angle loss utilizing triangle inequality to constrain the similarity angle of samples within triplets. More recently, Thomas $et~al.$~\cite{ANG} proposed a within-modality loss, which guides the semantically related images closer to each other. In this paper, we proposes a new metric learning strategy which can be applied to various models. The approach exploits the semantics within the visual modality and achieves better alignment between images and texts.

\begin{figure*}[t]
    \centering
    \includegraphics[width=17cm, trim=60 190 60 200]{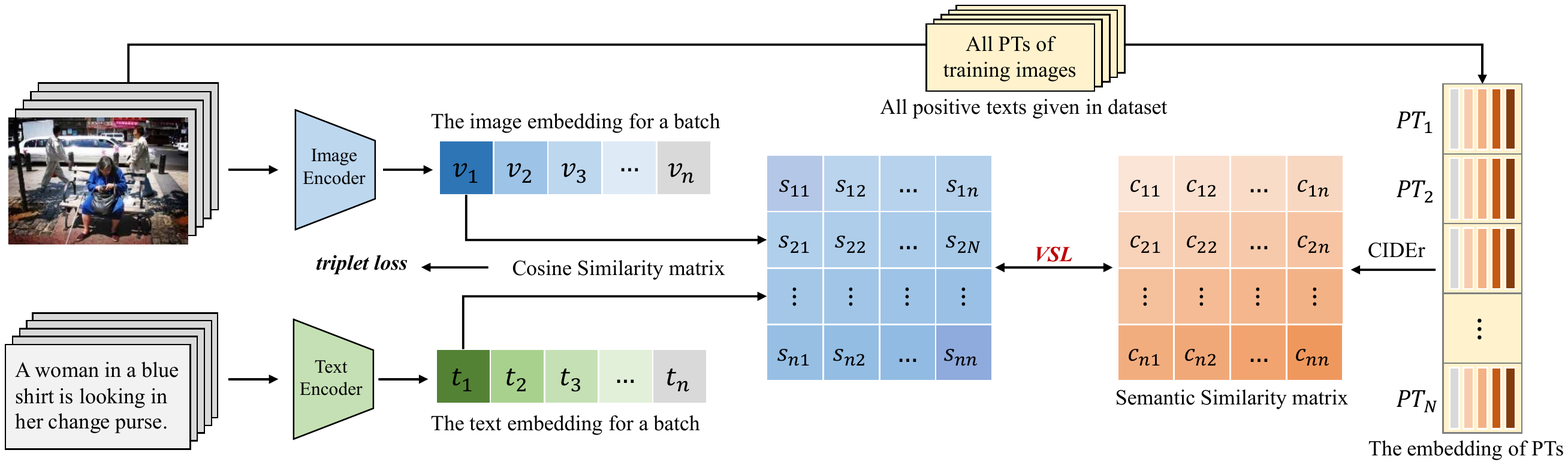}
    \caption{Illustration of the proposed Visual Semantic Loss (VSL) for image-text retrieval. Embeddings of image and text generated from a two-branch model are used to get the cosine similarity matrix. PTs of images are used to calculate CIDEr score and obtain a semantic similarity matrix among a mini-batch. The proposed VSL is a supplement to the model and allows the model to focus on the main content of the image by narrowing the difference between the two similarity matrices.}
    \label{f2}
\end{figure*}

\section{method}

In this section, we present \textbf{V}isual \textbf{S}emantic \textbf{L}oss (VSL). We first give the problem definition, and then explain the conceptual work and advantages of VSL in detail.

\subsection{Problem Definition}
Let there be a dataset $D=\{I,T\}$ of image-text pairs, where $I=\{v_1, v_2, ..., v_N\}$ denotes samples of the visual modality while $T=\{t_1, t_2, ..., t_N\}$ are samples of the textual modality. The notation $S=(s_{ij})_{n\times{n}}$, where $n$ is the batch size, represents the cosine similarity matrix predicted by the model. And the $s_{ij} = Sim(v_i, t_j)$ indicates the similarity of the $i$-th image with the $j$-th text as calculated by Eq.~\ref{1}.
\begin{equation}
Sim(v, t) = \frac{f_I(v)\cdot f_T(t)}{\Vert f_I(v)\Vert\cdot\Vert f_T(t)\Vert}
    \tag{1}\label{1}
\end{equation}
where $v\in{I}$ and $t\in{T}$. Additionally, $f_I(\cdot)$ denotes a visual encoder for mapping images into the common embedding space, and $f_T(\cdot)$ denotes a textual encoder for mapping texts. 

We employ the widely used triplet loss~\cite{FaceNet} as basic objective function (Eq.~\ref{2}). By constructing a triplet $(v,t,t^-)$ or $(t,v,v^-)$, where $v^-$ and $t^-$ are negative samples corresponding to the positive pair $(v,t)$, the method narrows the distance between positive pairs smaller than the distance between positive and negative samples.

\begin{equation}
 \begin{aligned}
        \mathcal{L}_{triplet} = &[m-Sim(v,t)+Sim(v^-,t)]_+\\
    +&[m-Sim(v,t)+Sim(v,t^-)]_+
 \end{aligned}
    \tag{2}\label{2}
\end{equation}
where $m$ is the margin parameter to prevent the anchor from having the same distance to the positive and negative samples. $Sim(\cdot)$ is the similarity as calculated by Eq.~\ref{1}.

Although triplet loss allows the model to learn alignment between different modalities, it cannot handle the challenge of fine-grained alignment caused by the mismatch between the text and the secondary content in an image. Therefore, we propose VSL to achieve fine-grained alignment by preserving the harmonization of visual and semantic similarity.

\subsection{Visual Semantic Loss}
The VSL is proposed to avoid mismatches caused by the secondary content of the image being more relevant to the text than the main content. Intuitively, it is employed to ensure the consistency of the intra-modal semantic similarity and inter-modal cosine similarity shown in Fig.~\ref{f2}. Concretely, when the annotated texts of image A and image B have high semantic similarity, their main contents will also be semantically similar due to the fact that people spontaneously focus on the main content when describing an image. Thus, the paired text of image A should be more semantically similar to image B than other images with unrelated main contents, but no more than image A. 

Therefore, we utilize the PTs (5 per image in MSCOCO) of images to represent the main semantics. Suppose $n$ is the batch size, then $C = (c_{ij})_{n\times{n}}$ denotes the semantic similarity matrix within visual modality. The $c_{ij}=K(v_i, v_j)$ is the similarity between the $i$-th and $j$-th image, where $K(\cdot)$ is reformulated from CIDEr~\cite{CIDEr}.
\begin{equation}
    K(v_i, v_j) = \frac{1}{Q^2}\sum_{p=1}^Q\sum_{q=1}^Q\frac{g(pt_i^p)\cdot g(pt_j^q)}{\Vert g(pt_i^p)\Vert\cdot \Vert g(pt_j^q)\Vert}
    \tag{3}\label{3}
\end{equation}
where $Q$ is the number of PTs per image, $pt_i^p$ is the $p$-th positive text of $v_i$, and $g(pt_i^p)$ is the vector of $pt_i^p$ computed through TFIDF~\cite{CIDEr}.

Because of the consistency of visual and semantics in the representation of the main content, the inter-modal cosine similarity matrix $S$ and the intra-modal semantic similarity matrix $C$ should also be converged. Accordingly, we design VSL to preserve the consistency of inter- and intra- modal similarity. To keep two similarity matrices, $S$ and $C$, in the same order of magnitude, we transform similarity matrices into ranking matrices using Eq.~\ref{4}. Suppose $SR = (sr_{ij})_{n\times{n}}$, where $sr_{ij}$ denotes the rank score of $s_{ij}$ in the $i$-th row of $S$, and so as $CR$. Inspired by~\cite{RankFunction, smoothAP}, we leverage the sigmoid function with a tunable parameter, written as $\Omega(\cdot)$, to reformulate the rank function as follows:
\begin{equation}
\begin{aligned}
    R(i, j, M) &= 1 + \sum_{k=1}^n\Omega(m_{ij} - m_{ik})\\
    \Omega(x) &= \frac{1}{1+exp(-x/\tau)}  
\end{aligned}
    \tag{4}\label{4}
\end{equation}
where $M = (m_{ij})_{n\times{n}}$ is a matrix to be ranked, $n$ is the batch size, and $\tau$ is a temperture parameter to control the slope of the sigmoid function. The ranking matrices $SR$ and $CR$ are obtained by $sr_{ij} = R(i,j,S)$ and $cr_{ij} = R(i,j,C)$. In this work, we set $\tau=0.001$ to approximate the ranking matrix calculated using the indicator function~\cite{RankFunction} and also avoid gradient disappearance~\cite{smoothAP}.

Eventually, the proposed visual semantic loss function is defined follows.
\begin{equation}
    \mathcal{L}_{vs} = 1-\frac{1}{n^2}\sum_{i=1}^n\sum_{j=1}^n\frac{Min\langle sr_{ij},cr_{ij}\rangle}{Max\langle sr_{ij},cr_{ij}\rangle}
    \tag{5}
\end{equation}
In summary, VSL is designed to narrow the difference between the two rank matrices, thus enhancing the representation of the main semantics of the image, which achieves a better alignment between text and image.
Section~\ref{setting}.

\begin{table*}[!h]
\centering
\caption{The results on the MSCOCO 1K and 5K. '*' means the baseline reproduced through public code.}
\label{t1}
\scalebox{1.1}{
\begin{tabular}{ccccccc|cccccc}
\hline
\multirow{3}{*}{Method} & \multicolumn{6}{c|}{MSCOCO 1K}                                                                & \multicolumn{6}{c}{MSCOCO 5K}                                                                 \\
                        & \multicolumn{3}{c}{Image to Text}             & \multicolumn{3}{c|}{Text to Image}            & \multicolumn{3}{c}{Image to Text}             & \multicolumn{3}{c}{Text to Image}             \\
                        & R@1           & R@5           & R@10          & R@1           & R@5           & R@10          & R@1           & R@5           & R@10          & R@1           & R@5           & R@10          \\ \hline
VSRN~\cite{VSRN}                    & 76.2          & 94.8          & 98.2          & 62.8          & 89.7          & 95.1          & 53.0            & 81.1          & 89.4          & 40.5          & 70.6          & 81.1          \\
IMRAM~\cite{IMRAM}                   & 76.7          & 95.6          & 98.5          & 61.7          & 89.1          & 95.0            & 53.7          & 83.2          & 91.0            & 39.7          & 69.1          & 79.8          \\
MMCA~\cite{MMCA}                    & 74.8          & 95.6          & 97.7          & 61.6          & 89.8          & 95.2          & 54.0            & 82.5          & 90.7          & 38.7          & 69.7          & 80.8          \\
NCR~\cite{NCR}                     & 78.7          & 95.8          & 98.5          & 63.3          & 90.4          & 95.8          & 58.2          & 84.2          & 91.5          & 41.7          & 71.0            & 81.3          \\
CMHF~\cite{CMHF}                    & 75.5          & 95.8          & 98.2          & 60.4          & 89.6          & 95.4          & 52.3          & 82.9          & 90.1          & 39.3          & 69.6          & 80.6          \\
CGMN~\cite{CGMN}                    & 76.8          & 95.4          & 98.3          & 63.8          & 90.7          & 95.7          & 53.4          & 81.3          & 89.6          & 41.2          & 71.9          & 82.4          \\
UARDA~\cite{UARDA}                   & 78.6          & 96.5          & \textbf{98.9} & 63.9          & 90.7          & \textbf{96.2}          & 56.2          & 83.8          & 91.3          & 40.6          & 69.5          & 80.9          \\ \hline
SGR*~\cite{SGRAF}                    & 77.1          & 95.9          & 98.4          & 62.5          & 89.5          & 95.3          & 56.0            & 82.9          & 91.2          & 40.4          & 70.0            & 80.6          \\
SGR+VSL (ours)            & 78.5          & 96.2          & 98.6          & 63.0            & 89.9          & 95.3          & 57.7          & 84.3          & 91.0          & 41.4          & 70.5          & 80.8            \\
SAF*~\cite{SGRAF}                    & 77.5          & 96.2          & 98.6          & 62.2          & 89.7          & 95.4          & 56.5          & 83.3          & 90.9          & 40.4          & 70.0          & 80.5          \\
SAF+VSL (ours)            & 78.3          & 96.0            & 98.6          & 63.0            & 89.9          & 95.3          & 56.2          & 84.4          & 91.3          & 41.4          & 70.4          & 81.0            \\
SGRAF*~\cite{SGRAF}                  & 79.3          & \textbf{96.7} & 98.7          & 63.8          & 90.4          & 95.7          & 58.5          & 84.5          & 92.1          & 42.1          & 71.4          & 81.8          \\
SGRAF+VSL (ours)          & \textbf{80.1} & 96.5          & 98.8          & \textbf{64.8} & \textbf{90.7} & 95.9 & \textbf{60.2} & \textbf{86.6} & \textbf{92.5} & \textbf{43.3} & \textbf{72.2} & \textbf{82.5} \\ \hline
\end{tabular}%
}
\end{table*}

\subsection{Optimization}
Through all the above descriptions, the final objective function is jointly optimized as Eq.~\ref{6}.
\begin{equation}
    \mathcal{L}=\alpha \mathcal{L}_{triplet} + \beta \mathcal{L}_{vs}
    \tag{6}\label{6}
\end{equation}
where $\alpha$ and $\beta$ are used to balance the contribution of $\mathcal{L}_{triplet}$ and $\mathcal{L}_{vs}$. Multiple experiments illustrate that $\alpha=1$ and $\beta=10$ is the best setting to train the model.
 
\section{Experiment}
\subsection{Experimental Setup}
\textbf{Datasets.} The proposed approach is evaluated on two benchmark datasets, MSCOCO~\cite{COCO} and Flickr30K~\cite{f30k}, composed of paired images and texts. The MSCOCO dataset contains 123,287 images, and each image has 5 annotated captions as positive texts. As in~\cite{SGRAF}, the dataset is split into 113,287 images for training, 5,000 images for validation, and 5,000 images for testing. Experimental results are reported on the 1K and 5K test sets. The Flickr30K dataset is composed of 31,783 images with 5 annotated texts per image. As in prior study~\cite{DeViSE}, 29,783 images are used for training, 1,000 images for validation, and 1,000 images for testing.

\noindent\textbf{Implementation Details.}\label{setting}
The VSL is a general learning strategy which could enable almost all existing cross-modal retrieval model to preserve semantic alignment between different modalities. To verify the effectiveness of the proposed approach, SGRAF~\cite{SGRAF} (containing SGR and SAF modules) is chosen as the base architecture, because it is the state-of-the-art model in image-text retrieval. As in~\cite{SGRAF}, the embedding dimension is set to 2048, the batch size is set to 128 and Adam optimizer is employed. The temperature parameter $\tau=0.001$ are utilized in Eq.~\ref{4}. For MSCOCO, the learning rate is set to 0.0003 for the first 10 epochs and 0.00003 for the next 15 epochs. For Flickr30K, the learning rate of SGR+VSL (SAF+VSL), is set to 0.0003 for the first 25 (10) steps and 0.00003 for the next 20 (15) steps.

\noindent\textbf{Evaluation Metrics.} The approach's performance is evaluated by widely used rank-based metric Recall@K (R@K) defined as the proportion of queries in which the ground truth is ranked in the top-K, where K=1, 5, 10. 

\subsection{Comparison with Prior Methods}
This section presents comparisons of the proposed approach on two datasets with the state-of-the-art methods. The benchmark methods are VSRN~\cite{VSRN}, MMCA~\cite{MMCA}, CMHF~\cite{CMHF} and CGMN~\cite{CGMN}, which use intra-modal information, and IMRAM~\cite{IMRAM}, NCR~\cite{NCR}, UARDA~\cite{UARDA} and SGRAF~\cite{SGRAF}, which do not.

\textbf{Results on MSCOCO.} The qualitative experimental results on MSCOCO 1K and 5K test sets are presented in Table~\ref{t1}. It is found that the three baselines with VSL all achieve better performance than before. 
For the results on MSCOCO 1K, compared with baselines, VSL improves R@1 for both SGR and SAF modules. Additionally, the proposed method outperformed other methods, and especially SGRAF+VSL achieves the best R@1 for both i2t (80.1\%) and t2i (64.8\%) tasks. Improvements are more significant for the MSCOCO 5K test set. Compared with baseline, SGRAF+VSL improves the R@1 by 1.7\% for i2t and 1.2\% for t2i. In general, the proposed VSL considerably outperforms all baselines demonstrating the effectiveness of the VSL.

\begin{table}[!h]
\centering
\caption{The results on the Flickr30K. “*” means the baseline reproduced through public code.}
\label{t2}
\resizebox{\columnwidth}{!}{%
\normalsize
\begin{tabular}{ccccccc}
\hline
\multirow{3}{*}{Method} & \multicolumn{6}{c}{Flickr30K}                                                                  \\
                        & \multicolumn{3}{c}{Image to Text}             & \multicolumn{3}{c}{Text to Image}             \\
                        & R@1           & R@5           & R@10          & R@1           & R@5           & R@10          \\ \hline
VSRN~\cite{VSRN}                    & 71.3          & 90.6          & 96.0            & 54.7          & 81.8          & 88.2          \\
IMRAM~\cite{IMRAM}                   & 74.1          & 93.0            & 96.6          & 53.9          & 79.4          & 87.2          \\
MMCA~\cite{MMCA}                    & 74.2          & 92.8          & 96.4          & 54.8          & 81.4          & 87.8          \\
NCR~\cite{NCR}                     & 77.3          & 94.0            & 97.5          & 59.6          & 84.4          & 89.9          \\
CMHF~\cite{CMHF}                    & 63.6          & 88.6          & 94.0            & 45.4          & 76.6          & 85.0            \\
CGMN~\cite{CGMN}                    & 77.9          & 93.8          & 96.8          & 59.9          & \textbf{85.1} & \textbf{90.6} \\
UARDA~\cite{UARDA}                   & 77.8          & 95.0            & 97.6          & 57.8          & 82.9          & 89.2          \\ \hline
SGR*~\cite{SGRAF}                    & 75.4          & 93.4          & 96.9          & 56.2          & 81.4          & 86.2          \\
SGR+VSL (ours)            & 75.7          & 93.5          & 96.5          & 56.5          & 80.9          & 85.9          \\
SAF*~\cite{SGRAF}                    & 72.9          & 92.7          & 95.6          & 55.6          & 81.5          & 87.8          \\
SAF+VSL (ours)           & 75.9          & 93.9          & 97.5          & 57.9          & 82.7          & 88.9          \\
SGRAF*~\cite{SGRAF}                  & 77.1          & 94.3          & 97.3          & 57.7          & 83.0            & 88.6          \\
SGRAF+VSL (ours)          & \textbf{79.5} & \textbf{95.3} & \textbf{97.9} & \textbf{60.2} & 84.3          & 89.4          \\ \hline
\end{tabular}%
}
\end{table}
\textbf{Results on Flickr30K.} The results on Flickr30K are reported in Table~\ref{t2}. 
It is observed that models with VSL show significant improvement compared to all baselines, especially SGRAF+VSL which achieves the best R@1 79.5\% (2.4\%$\uparrow$) for i2t and 60.2\% (2.5\%$\uparrow$) for t2i. The results indicate that the VSL has stable improvements of the performance of models.

\subsection{Ablation Studies}

In this section, results from extensive ablation studies are reported on MSCOCO to verify the contribution of VSL. Ablation experiments on two aspects are conducted, one is the effect of data diversity and the other is the effect of semantic-based loss function within the visual and textual modality. And the analysis experiments of parameter $\alpha$ and $\beta$ in Eq.~\ref{4} are also done to find the best setting.

\textbf{Data diversity.} To explore the impact of data diversity on retrieval, we design experiments with batchsize = 32, 64, and 128, since the data diversity within a batch is positively correlated with the batch size. The results are shown in Table~\ref{t3}. It can be found that, for almost all batchsize settings, the model with VSL outperforms the baseline on R@1. The improvement is greater for a larger batch size, and the best performance is obtained for batchsize = 128, where R@1 is improved by 1.4\% for i2t and 0.5\% for t2i. With the batchsize = 32 or 64, there is only slight improvement on R@1 or even a drop. The results demonstrate that the effectiveness of the VSL is positively correlated with data diversity. When the training data is well-diversified, the model can capture abundant semantic correspondences of images through VSL, and thus focus more on the main semantics of the images.
\begin{table}[!h]
\centering
\caption{The ablation study of batch size on the MSCOCO 1K. '*' means the baseline reproduced through public code.}
\label{t3}
\resizebox{\columnwidth}{!}{%
\normalsize
\begin{tabular}{lccccccc}
\hline
\multicolumn{1}{c}{}                         &                             & \multicolumn{6}{c}{MSCOCO 1K}                                                                           \\
\multicolumn{1}{c}{}                         &                             & \multicolumn{3}{c}{Image to Text}                  & \multicolumn{3}{c}{Text to Image}                  \\
\multicolumn{1}{c}{\multirow{-3}{*}{Method}} & \multirow{-3}{*}{Batchsize} & R@1                                  & R@5  & R@10 & R@1                                  & R@5  & R@10 \\ \hline
SGR*                                         & 32                          & 75.3                                 & 95.6 & 98.4 & 60.5                                 & 88.9 & 95.1 \\
\multicolumn{1}{r}{+VS}                      & 32                          & {\color[HTML]{32CB00} \textbf{75.1}} & 95.1 & 98.1 & \textbf{60.5} & 89.0 & 95.0 \\ \hline
SGR*                                         & 64                          & 76.5                                 & 95.5 & 98.2 & 61.8                                 & 89.4 & 95.3 \\
\multicolumn{1}{r}{+VS}                      & 64                          & {\color[HTML]{FE0000} \textbf{76.8}} & 95.7 & 98.5 & {\color[HTML]{FE0000} \textbf{62.0}} & 89.6 & 94.9 \\ \hline
SGR*                                         & 128                         & 77.1                                 & 95.9 & 98.4 & 62.5                                 & 89.5 & 95.3 \\
\multicolumn{1}{r}{+VS}                      & 128                         & {\color[HTML]{FE0000} \textbf{78.5}} & 96.2 & 98.6 & {\color[HTML]{FE0000} \textbf{63.0}} & 89.9 & 95.3 \\ \hline
\end{tabular}%
}
\end{table}

\begin{figure*}[!h]
    \centering
    \includegraphics[width=17cm, trim= 35 180 35 180]{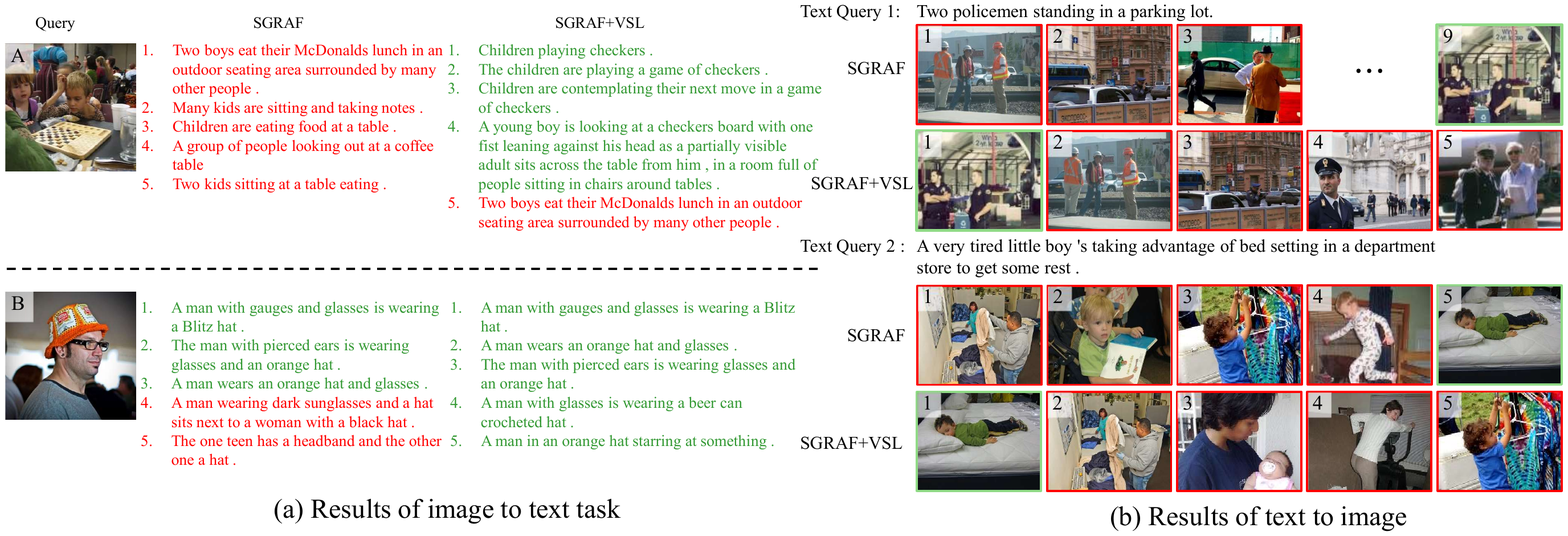}
    \caption{Visualization of image-text retrieval results on Flickr30K. (a) shows the ranking results for the i2t task. Positive texts are in green while the negative texts are in red. (b) shows the ranking results for the t2i task. Positive images are in green boxes while negative ones are in red boxes.}
    \label{f3}
\end{figure*}

\textbf{Semantic similarity within the visual and textual modality.} To demonstrate that the main semantics of images, which is ignored before, plays a major role in enhancing visual and textual alignment, a intra-modal semantic similarity loss function within the text modality is symmetrically designed shown as  Eq.~\ref{7}, marked as $\mathcal{L}_{ts}$.

\begin{equation}
    \begin{aligned}
     \mathcal{L}_{ts} &= 1-\frac{1}{n^2}\sum_{i=1}^n\sum_{j=1}^n\frac{Min\langle sr'_{ij},tcr_{ij}\rangle}{Max\langle sr'_{ij},tcr_{ij}\rangle}\\
    \mathcal{L} &= \alpha{L_{triplet}}+\beta{(L_{vs}+L_{ts})}
    \end{aligned}
    \tag{7}\label{7} 
\end{equation}
where $sr'_{ij}=R(i,j,S')$ is the rank score of $s'_{ij}$ calculated by Eq.~\ref{4}. The $S'$ is the transpose of similarity matrix $S$. The $tcr_{ij}=R(i,j,TC)$ is the rank score of textual semantic similarity $tc_{ij}$ in $i^{th}$ row, where $TC=(tc_{ij})_{n\times{n}}$ is the semantic similarity matrix within textual modality calculated by CIDEr~\cite{CIDEr} among a mini-batch. And $\mathcal{L}$ is the final objective function for ablation study with $\alpha=1$ and $\beta=10$.

\begin{table}[!h]
\centering
\caption{The ablation study of two intra-modal semantic loss functions on the MSCOCO 5K.}
\label{t4}
\resizebox{\columnwidth}{!}{%
\normalsize
\begin{tabular}{ccccccccc}
\hline
\multirow{3}{*}{Method} & \multirow{3}{*}{$\mathcal{L}_{vs}$} & \multirow{3}{*}{$\mathcal{L}_{ts}$} & \multicolumn{6}{c}{MSCOCO 5K}                                                                 \\
                        &                     &                     & \multicolumn{3}{c}{Image to Text}             & \multicolumn{3}{c}{Text to Image}             \\
                        &                     &                     & R@1           & R@5           & R@10          & R@1           & R@5           & R@10          \\ \hline
SGRAF*                  &                     &                     & 58.5          & 84.5          & 92.1          & 41.6          & 70.9          & 81.5          \\
SGRAF+            & \Checkmark                   & \Checkmark                   & 59.5          & 86.2          & \textbf{92.5} & 43.1          & 72.1          & 82.1          \\
SGRAF+            & \Checkmark                   &                     & \textbf{60.2} & \textbf{86.6} & \textbf{92.5} & \textbf{43.3} & \textbf{72.2} & \textbf{82.5} \\ \hline
\end{tabular}%
}
\end{table}
The impacts of $\mathcal{L}_{vs}$ and $\mathcal{L}_{ts}$ are shown in Table.~\ref{t4}. Compared with baseline, the model with both $\mathcal{L}_{vs}$ and $\mathcal{L}_{ts}$ also achieves an improvement on R@1. However, compared with the model using only $\mathcal{L}_{vs}$, the performance of model with $\mathcal{L}_{vs}$ and $\mathcal{L}_{ts}$ is worse on R@1, dropping by 0.7\% for i2t and 0.2\% for t2i. The results show that the main semantic of the visual modality indeed enhances the alignment of the image and text. And $\mathcal{L}_{vs}$ has crucial effect on constraining the model to preserving the main semantics of images and texts.

\textbf{Analysis experiments of parameters.} The parameters $\alpha$ and $\beta$ in Eq.~\ref{4} are used to balance the contribution of triplet loss and VSL. To balance the individual contribution of each portion of loss, we conduct analysis experiments of $\alpha$ and $\beta$ to find the most reasonable experiment setting. The results of analysis experiments are shown in Table.~\ref{t5}. It can be observed that the performance of SAF+VSL reaches the peak when $\alpha=1$ and $\beta=10$. Since the VSL takes on the primary role of semantic alignment during training, whether the value of $\beta$ is too large or too small, the R@1 drops significantly, which indicates the reasonableness of the experiment setting.


\begin{table}[H]
\centering
\caption{The analysis experiments of parameter $\alpha$ and $\beta$.}
\label{t5}
\resizebox{\columnwidth}{!}{%
\begin{tabular}{ccccccccc}
\hline
\multirow{3}{*}{Approach} & \multicolumn{2}{c}{\multirow{2}{*}{Parameters}} & \multicolumn{6}{c}{MSCOCO 1K}                                                             \\
                          & \multicolumn{2}{c}{}                            & \multicolumn{3}{c}{Image to Text}           & \multicolumn{3}{c}{Text to Image}           \\
                          & $\alpha$                      & $\beta$                      & R@1           & R@5         & R@10          & R@1         & R@5           & R@10          \\ \hline
SAF+VSL                   & 1                      & 6                      & 74.0          & 93.6        & 96.9          & 56.5        & 83.0          & 88.6          \\
\textbf{SAF+VSL}          & 1                      & 10                     & \textbf{78.3} & \textbf{96.0} & \textbf{98.6} & \textbf{63.0} & \textbf{89.9} & \textbf{95.3} \\
SAF+VSL                   & 1                      & 15                     & 72.1          & 94.1        & 97.6          & 60.9        & 89.0          & 94.9          \\
SAF+VSL                   & 1                      & 20                     & 73.2          & 93.7        & 96.6          & 60.7        & 87.8          & 93.5          \\
SAF+VSL                   & 1                      & 25                     & 73.4          & 93.9        & 96.6          & 60.9        & 88.0            & 93.5          \\ \hline
\end{tabular}%
}
\end{table}
\subsection{Qualitative Analysis}
To further verify the performance of the approach, several representative images and texts are selected in the Flickr30K test set and the retrieval results are visualized. The text retrieval results are shown in Fig.~\ref{f3}(a). For the hard sample image A in text retrieval, the baseline failed to retrieve any of the positive texts in the top-5, while the proposed approach correctly matched most of the ground truth. And for the normal sample image B, the proposed VSL supplements the rest of the positive texts which are mismatched by the baseline. The image retrieval results are shown in Fig.~\ref{f3}(b). For the hard sample text query 1, the baseline cannot match the positive image until rank ninth while ours retrieves the true image at top-1. And for the normal sample text query 2, the positive image is ranked fifth in results from~\cite{SGRAF}, while ours matches the correct one at top-1.
\section{Conclusion}
This paper propose a novel visual semantic loss function (VSL), which utilizes the intra-modal semantics to constrain embedding learning. With the contribution of CIDEr, the proposed approach mines semantic information within the visual modality to assist the existing models in capturing the main content of the image, which enhances the alignment of the text with the main content of the image. Experimental results on two widely used datasets indicate that the VSL is an effective constraint that can be applied to improve the performance of the image-text retrieval.

\bibliographystyle{IEEEbib}
\bibliography{ref.bib}

\begin{thebibliography}{10}

\bibitem{SGRAF}
Haiwen Diao, Ying Zhang, Lin Ma, and Huchuan Lu,
\newblock ``Similarity reasoning and filtration for image-text matching,''
\newblock in {\em AAAI}, 2021, pp. 1218--1226.

\bibitem{IMRAM}
Hui Chen, Guiguang Ding, Xudong Liu, Zijia Lin, Ji~Liu, and Jungong Han,
\newblock ``{IMRAM}: Iterative matching with recurrent attention memory for
  cross-modal image-text retrieval,''
\newblock in {\em CVPR}, 2020, pp. 12655--12663.

\bibitem{MMCA}
Xi~Wei, Tianzhu Zhang, Yan Li, Yongdong Zhang, and Feng Wu,
\newblock ``Multi-modality cross attention network for image and sentence
  matching,''
\newblock in {\em CVPR}, 2020, pp. 10941--10950.

\bibitem{ILSA}
Kai Wang, Yifan Wang, Xing Xu, Zuo Cao, and Xunliang Cai,
\newblock ``Instance-level semantic alignment for zero-shot cross-modal
  retrieval,''
\newblock in {\em ICME}, 2022, pp. 1--6.

\bibitem{CMHF}
Xing Xu, Yifan Wang, Yixuan He, Yang Yang, Alan Hanjalic, and Heng~Tao Shen,
\newblock ``Cross-modal hybrid feature fusion for image-sentence matching,''
\newblock {\em TOMCCAP}, vol. 17, no. 4, pp. 1--23, 2021.

\bibitem{DAA}
Hao Li, Jingkuan Song, Lianli Gao, Pengpeng Zeng, Haonan Zhang, and Gongfu Li,
\newblock ``A differentiable semantic metric approximation in probabilistic
  embedding for cross-modal retrieval,''
\newblock {\em NeurIPS}, pp. 11934--11946, 2022.

\bibitem{CIDEr}
Ramakrishna Vedantam, C.~Lawrence Zitnick, and Devi Parikh,
\newblock ``Cider: Consensus-based image description evaluation,''
\newblock in {\em CVPR}, 2015, pp. 4566--4575.

\bibitem{COCO}
Tsung-Yi Lin, Michael Maire, Serge Belongie, James Hays, Pietro Perona, Deva
  Ramanan, Piotr Doll{\'a}r, and C.~Lawrence Zitnick,
\newblock ``Microsoft coco: Common objects in context,''
\newblock in {\em ECCV}, 2014, pp. 740--755.

\bibitem{f30k}
Peter Young, Alice Lai, Micah Hodosh, and Julia Hockenmaier,
\newblock ``From image descriptions to visual denotations: New similarity
  metrics for semantic inference over event descriptions,''
\newblock {\em TACL}, vol. 2, pp. 67--78, 2014.

\bibitem{Semi}
Thomas~N. Kipf and Max Welling,
\newblock ``Semi-supervised classification with graph convolutional networks,''
\newblock in {\em ICLR}, 2017, pp. 1--14.

\bibitem{VSRN}
Kunpeng Li, Yulun Zhang, Kai Li, Yuanyuan Li, and Yun Fu,
\newblock ``Visual semantic reasoning for image-text matching,''
\newblock in {\em ICCV}, 2019, pp. 4654--4662.

\bibitem{C3CMR}
Junsheng Wang, Tiantian Gong, Zhixiong Zeng, Changchang Sun, and Yan Yan,
\newblock ``C3cmr: Cross-modality cross-instance contrastive learning for
  cross-media retrieval,''
\newblock in {\em ACM MM}, 2022, pp. 4300--4308.

\bibitem{ALBEF}
Junnan Li, Ramprasaath Selvaraju, Akhilesh Gotmare, Shafiq Joty, Caiming Xiong,
  and Steven Chu~Hong Hoi,
\newblock ``Align before fuse: Vision and language representation learning with
  momentum distillation,''
\newblock {\em {NeurIPS}}, vol. 34, pp. 9694--9705, 2021.

\bibitem{ABGR}
Xian Zhong, Zhengwei Yang, Mang Ye, Wenxin Huang, Jingling Yuan, and Chia{-}Wen
  Lin,
\newblock ``Auxiliary bi-level graph representation for cross-modal image-text
  retrieval,''
\newblock in {\em ICME}, 2021, pp. 1--6.

\bibitem{FaceNet}
Florian Schroff, Dmitry Kalenichenko, and James Philbin,
\newblock ``{FaceNet}: A unified embedding for face recognition and
  clustering,''
\newblock in {\em CVPR}, 2015, pp. 815--823.

\bibitem{N-pair}
Kihyuk Sohn,
\newblock ``Improved deep metric learning with multi-class n-pair loss
  objective,''
\newblock in {\em NeurIPS}, 2016, p. 1857–1865.

\bibitem{angle}
Jian Wang, Feng Zhou, Shilei Wen, Xiao Liu, and Yuanqing Lin,
\newblock ``Deep metric learning with angular loss,''
\newblock in {\em {ICCV}}, 2017.

\bibitem{ANG}
Christopher Thomas and Adriana Kovashka,
\newblock ``Preserving semantic neighborhoods for robust cross-modal
  retrieval,''
\newblock in {\em {ECCV}}, 2020.

\bibitem{RankFunction}
Tao Qin, Tie-Yan Liu, and Hang Li,
\newblock ``A general approximation framework for direct optimization of
  information retrieval measures,''
\newblock {\em Inf. Retr.}, vol. 13, no. 4, pp. 375--397, 2010.

\bibitem{smoothAP}
Andrew Brown, Weidi Xie, Vicky Kalogeiton, and Andrew Zisserman,
\newblock ``Smooth-ap: Smoothing the path towards large-scale image
  retrieval,''
\newblock in {\em {ECCV}}, 2021, pp. 677--694.

\bibitem{NCR}
Zhenyu Huang, Guocheng Niu, Xiao Liu, Wenbiao Ding, Xinyan Xiao, Hua Wu, and
  Xi~Peng,
\newblock ``Learning with noisy correspondence for cross-modal matching,''
\newblock in {\em NeurIPS}, 2021, pp. 29406--29419.

\bibitem{CGMN}
Yuhao Cheng, Xiaoguang Zhu, Jiuchao Qian, Fei Wen, and Peilin Liu,
\newblock ``Cross-modal graph matching network for image-text retrieval,''
\newblock {\em TOMCCAP}, vol. 18, no. 4, pp. 1--23, 2022.

\bibitem{UARDA}
Kun Zhang, Zhendong Mao, Anan Liu, and Yongdong Zhang,
\newblock ``Unified adaptive relevance distinguishable attention network for
  image-text matching,''
\newblock {\em TMM}, vol. 14, no. 8, pp. 1--14, 2022.

\bibitem{DeViSE}
Andrea Frome, Gregory~S. Corrado, Jonathon Shlens, Samy Bengio, Jeffrey Dean,
  Marc'Aurelio Ranzato, and Tom{\'{a}}s Mikolov,
\newblock ``Devise: {A} deep visual-semantic embedding model,''
\newblock in {\em {NeurIPS}}, 2013, pp. 1--14.

\end{thebibliography}

\end{document}